\documentclass[journal]{IEEEtran}
\usepackage{multirow}
\usepackage{amsfonts}

\usepackage{booktabs}

\usepackage{makecell}
\usepackage{threeparttable}

\usepackage{float}

\newtheorem{remark}{Remark}
\usepackage[numbers,sort&compress]{natbib}
\usepackage{colortbl}
\usepackage[dvipsnames]{xcolor}

\usepackage{stfloats}
\usepackage{ifpdf}

   \usepackage{graphicx}
   \graphicspath{{./figures/}}
\usepackage{amsmath}
\usepackage{algorithmicx}

\usepackage{algorithm}  
\usepackage{algpseudocode}  
\usepackage{amsmath}  

\usepackage{array}

\ifCLASSOPTIONcompsoc
\usepackage[caption=false,font=normalsize,labelfont=sf,textfont=sf]{subfig}
\else
\usepackage[caption=false,font=footnotesize]{subfig}
\usepackage{fixltx2e}

\usepackage{stfloats}
\usepackage{url}

\begin{document}
%
\title{Interpretable Deep Reinforcement Learning for Optimizing Heterogeneous Energy Storage Systems}
%
%

\author{Luolin~Xiong,~\IEEEmembership{Graduate~Student~Member,~IEEE},~Yang~Tang,~\IEEEmembership{Senior~Member,~IEEE},\\~Chensheng~Liu,~\IEEEmembership{Member,~IEEE},~Shuai~Mao,~Ke~Meng,~\IEEEmembership{Senior~Member,~IEEE},\\~Zhaoyang~Dong,~\IEEEmembership{Fellow,~IEEE},~Feng~Qian,~\IEEEmembership{Senior~Member,~IEEE}
\thanks{This work was supported by National Natural Science Foundation of China (61988101, 62293502, 62293504), Fundamental Research Funds for the Central Universities. \emph{(Corresponding author: Yang Tang, Feng Qian.)}}
\thanks{Luolin Xiong, Yang Tang, Chensheng Liu and Feng Qian are with the Key Laboratory of Smart Manufacturing in Energy Chemical Process, Ministry of Education, and the Engineering Research Center of Process System Engineering, Ministry of Education, East China University of Science and Technology, Shanghai 200237, China (e-mails: xiongluolin@gmail.com, tangtany@gmail.com, cliu@ecust.edu.cn, fqian@ecust.edu.cn).}
\thanks{Shuai Mao is with the Department of Electrical Engineering, Nantong University, Nantong 226019, China (e-mail: mshecust@163.com).}
\thanks{Ke Meng is with the School of Electrical Engineering and Telecommunications, University of New South	Wales, Sydney NSW 2052, Australia (e-mail: kemeng@ieee.org).}
\thanks{Zhaoyang Dong is with the School of Electrical and Electronics Engineering, Nanyang Technological University, 50 Nanyang Avenue 639798, Singapore (e-mail: zydong@ieee.org).}}

%
%

\markboth{IEEE TRANSACTIONS ON CIRCUITS AND SYSTEMS—I: REGULAR PAPERS}
{Xiong \MakeLowercase{\textit{et al.}}: Bare Demo of IEEEtran.cls for IEEE Journals}
%
\newcommand{\upcite}[1]{\textsuperscript{\textsuperscript{\cite{#1}}}}



\maketitle

\begin{abstract}
Energy storage systems (ESS) are pivotal component in the energy market, serving as both energy suppliers and consumers. ESS operators can reap benefits from energy arbitrage by optimizing operations of storage equipment. To further enhance ESS flexibility within the energy market and improve renewable energy utilization, a heterogeneous photovoltaic-ESS (PV-ESS) is proposed, which leverages the unique characteristics of battery energy storage (BES) and hydrogen energy storage (HES). For scheduling tasks of the heterogeneous PV-ESS, cost description plays a crucial role in guiding operator's strategies to maximize benefits. We develop a comprehensive cost function that takes into account degradation, capital, and operation/maintenance costs to reflect real-world scenarios. Moreover, while numerous methods excel in optimizing ESS energy arbitrage, they often rely on black-box models with opaque decision-making processes, limiting practical applicability. To overcome this limitation and enable transparent scheduling strategies, a prototype-based policy network with inherent interpretability is introduced. This network employs human-designed prototypes to guide decision-making by comparing similarities between prototypical situations and encountered situations, which allows for naturally explained scheduling strategies. Comparative results across four distinct cases underscore the effectiveness and practicality of our proposed pre-hoc interpretable optimization method when contrasted with black-box models.
\end{abstract}


%
\IEEEpeerreviewmaketitle

\section{Introduction}
%
%
%
%
\IEEEPARstart{A}{s} one of the significant resource, energy storage system (ESS), characterized by their flexibility, are extensively integrated into power systems, and contribute to carbon emission reduction \cite{9248603, 10081065, 123456, 9310351}. Flexible ESS serves a dual role in the energy market, functioning both as an energy supplier and consumer \cite{5371839}. One noteworthy application lies in its capacity to profit from participation in the energy market through energy arbitrage \cite{7892020}. Simultaneously, ESS can also support the integration of random renewable energy sources into the energy market, fostering competition with conventional non-renewable energy producers \cite{5641629, 6423224, 9935564}. Thoughtful scheduling of ESS operations can enhance the profitability of ESS operators and maximize the utilization of renewable energy resources, thereby mitigating the inherent uncertainties associated with renewable energy sources \cite{9789478}.

Many research efforts have been devoted to various ESS, such as battery energy storage (BES), hydrogen energy storage (HES), compressed air energy storage, and pumped hydro energy storage \cite{9782559, 7018991, 9857842, 9999551}. These ESS variants exhibit diverse dynamic characteristics. For instance, lithium batteries offer rapid response capabilities, enabling swift charging and discharging, whereas hydrogen and pumped hydro energy storage systems require more extended response times \cite{9782559}. Furthermore, lithium batteries have constrained capacities, especially when compared to compressed air energy storage and pumped hydro energy storage, which are better suited for large-scale applications \cite{7018991}. Hydrogen energy storage systems excel in energy density and storage duration \cite{9857842, 9999551}. Given the flexibility and variety within ESS, numerous studies have explored the combination of photovoltaic (PV) power stations with ESS to enhance overall energy efficiency \cite{9782559, 9310351}. In contrast to prior configurations involving PV-battery storage systems and PV-compressed air energy storage systems, we propose a unique combination of the PV system with both BES and HES as a PV-ESS, which leverages the distinctive characteristics of these heterogeneous energy storage systems, and further augment the revenue potential for the PV-ESS operator.

To treat a PV-ESS as an entity within power systems and optimize the economic profitability for the PV-ESS operator, a critical step involves the development of a realistic cost function. Significant efforts have been dedicated to describing the costs associated with various types of energy storage systems \cite{8805438, 9999506, 10039495, 9214988}. For instance, the degradation cost of BES has garnered considerable attention \cite{8805438, 9999506}. Factors such as depth of discharge (DoD), discharge rate, and state of charge (SoC) are recognized as pivotal in influencing battery degradation. \cite{8805438} has introduced a BES cost model, accounting for degradation cost based on DoD and discharge rate, which is applicable to conventional electrochemical batteries. \cite{9999506} has captured the intricacies of battery degradation mechanisms and presented a battery degradation model that considers both DoD and SoC. Additionally, the capital cost and the operation/maintenance cost hold substantial importance, particularly for large-scale energy storage equipment. \cite{9399252} has highlighted the impact of factors such as start-up/shut-down cycles, rapid transients in operational conditions, and the number of working hours on electrolyzers and fuel cells within HES systems. However, different from traditional homogeneous ESS, our proposed heterogeneous ESS poses a unique challenge in comprehensively characterizing various cost components. In this paper, we establish a comprehensive cost function that simultaneously incorporates degradation, capital, as well as operation/maintenance costs to mirror real-world scenarios.

On the other hand, due to the remarkable flexibility of ESS and the inherent uncertainty associated with PV power generation, the development of optimal scheduling strategies for PV-ESS has garnered significant research attention, particularly with the well-defined cost function \cite{9844280, 9925086, 9727092, 10064640}. Traditional optimization approaches for solving the scheduling problem have focused on mathematical programming and heuristic techniques \cite{9329108, 9999506, 9844159, li2018operation}. For instance, \cite{9999506} has implemented a mixed integer linear programming algorithm to address a day-ahead economic scheduling problem of BES systems, where a one-dimensional linearization technique has been used to linearize the two-variable function, reducing computational complexity without sacrificing accuracy. In \cite{9844159}, particle swarm optimization has been employed to obtain an optimal operation schedule for the ESS. Despite notable advancements in these traditional optimization methods, these approaches often hinge on precise models or assumptions about the distribution of random variables such as the PV power generation, which limits their applicability.

Recent advancements in artificial intelligence (AI) have led to the popularity of optimization methods based on reinforcement learning, which are particularly attractive for their independence on precise system models and strong performance, especially in uncertain systems \cite{9684239, 10083058, xiong2023home} and multi-agent systems \cite{6261564, wang2022cooperative}. In \cite{9310351}, a proximal policy optimization (PPO) based deep reinforcement learning (DRL) method has been employed to address capacity scheduling in PV-battery storage systems. This approach has demonstrated adaptability to uncertain market signals as well as PV generation profiles. Meanwhile, \cite{9782559} has introduced a model-free DRL technique for optimizing energy arbitrage, utilizing a hybrid model to forecast intermittent PV power generation. Nevertheless, the practical application of these AI-powered methods is somewhat constrained due to their opaque decision-making processes.

It has been widely recognized that interpretability is a crucial factor to enhance the practical applicability of reinforcement learning. A few works have explored the interpretability of reinforcement learning when applied to forecasting/optimization problems in energy system \cite{10068282, 9750856}. \cite{10068282} has provided the post-hoc interpretability for the policy network, employing the Shapley value to uncover the significance of various input features in the decision-making process. However, it's worth noting that this post-hoc explanation method is primarily retrospective, offering insights about the black-box model after the decision-making process. Consequently, it does not empower operators to understand the agent's decision-making process. 

Building upon the aforementioned motivation, this paper introduces an inherently interpretable DRL algorithm with pre-hoc interpretability tailored for addressing the heterogeneous PV-ESS scheduling problem. This pre-hoc explanation method relies on intuitive human-designed prototypes to guide the decision-making process by comparing similarities between prototypical situations and encountered situations \cite{kenny2022towards}. To achieve this, a prototype-based policy network is trained and integrated with a pre-trained agent, which can shorten the gap between the well-performed black-box model and the interpretable strategies. Our main contributions are summarized as follows:

\begin{itemize}
\item A heterogeneous PV-ESS is proposed to leverage the distinctive characteristics of BES and HES, thereby enhancing flexibility within the energy market. The scheduling problem of this PV-ESS is formulated as a Markov Decision Process (MDP), with the primary objective of maximizing benefits of the operator through energy arbitrage. Compared with homogeneous ESS in \cite{9782559, 9310351}, our approach incorporates a comprehensive cost function for the heterogeneous PV-ESS accounting for degradation, capital, as well as operation and maintenance costs to provide more realistic and practical guidance for operator decision-making.
\end{itemize}

\begin{itemize}
\item An interpretable DRL method is developed to provide pre-hoc interpretability for agent's decision-making process. Different from post-hoc explanation methods outlined in \cite{10068282}, our approach involves training a prototype-based policy network, which enables the guidance of decision-making by assessing similarities between human-defined prototypical situations and encountered situations, thereby rendering the decision-making process transparent. Significantly, this method reveals the correlation between the decisions made by the agent and the comprehensible decisions made by human.
\end{itemize}

\begin{itemize}
\item A comprehensive assessment across four cases, each featuring different PV-ESS configurations, is conducted. The comparative results illustrate the effectiveness and applicability of the proposed pre-hoc interpretable DRL method compared with black-box models. Furthermore, we evaluate the revenue of the PV-ESS operator, considering various scenarios with heterogeneous energy storage devices, and investigate the impact of the learning rate on convergence and optimization.
\end{itemize}

The remainder of this paper is structured as follows: Section II provides a detailed formulation of the scheduling problem for the heterogeneous PV-ESS. Section III outlines our proposed interpretable DRL method, featuring a prototype-based policy network. We present the simulation results in Section IV, and finally, we conclude the paper in Section V.

\section{Problem Formulation}
\begin{figure}[htb]	
	\centering	
	\includegraphics[scale=1, width=3.5in]{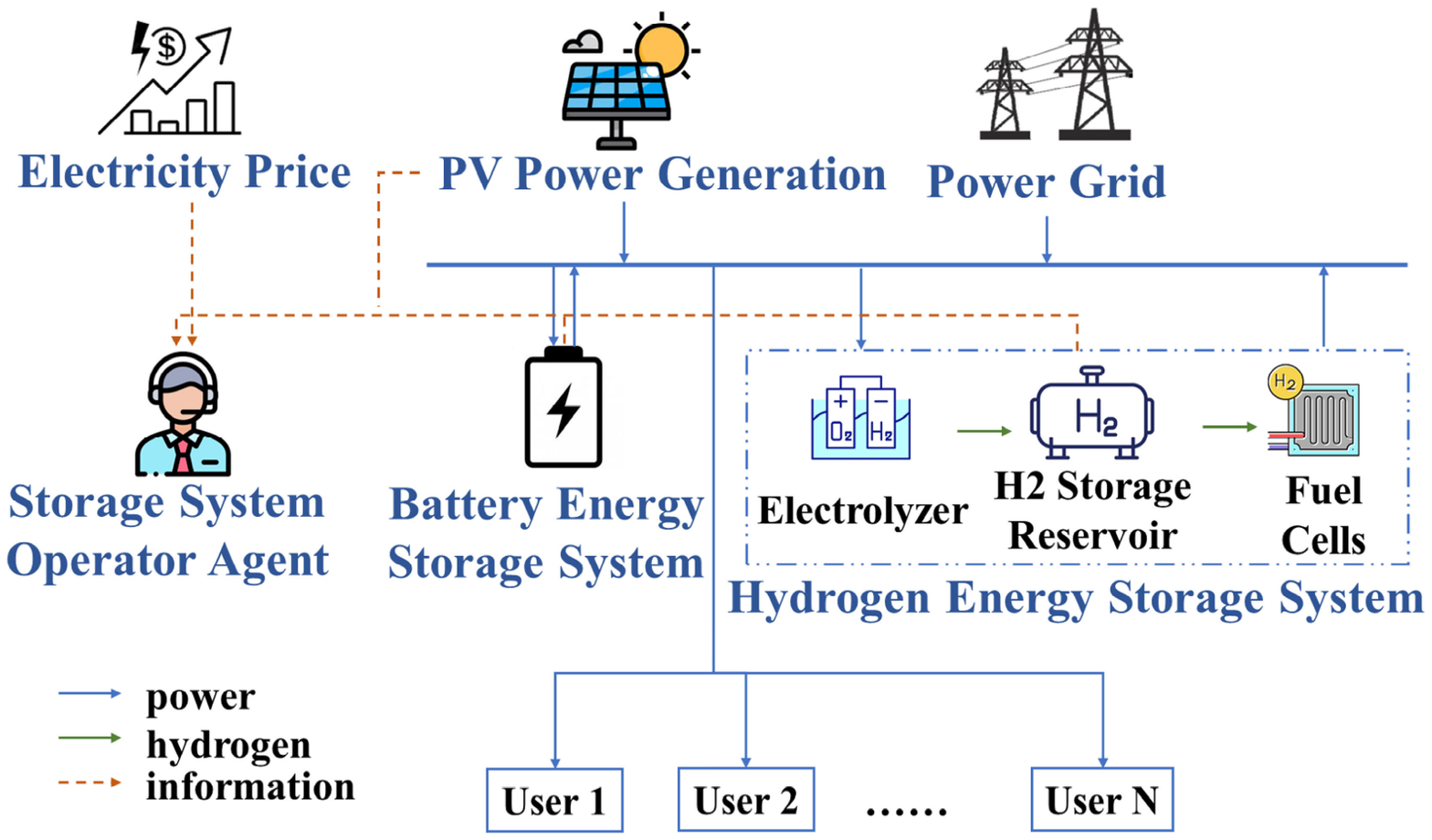}	
	\caption{Framework of the energy market with a heterogeneous PV-ESS.}	
	\label{Framework of energy market with heterogeneous ESS}
\end{figure}

As depicted in Fig. \ref{Framework of energy market with heterogeneous ESS}, our study focuses on energy arbitrage through the coordinated operations of the heterogeneous PV-ESS. The goal is to maximize the revenue of the PV-ESS operator by actively participating in electricity markets. In this section, we introduce dynamic models and the cost function of the heterogeneous PV-ESS, consisting of BES and HES. Subsequently, accounting for dynamic market prices and uncertainties associated with PV power generation, we formulate the operation scheduling task of the heterogeneous PV-ESS as a MDP. Within this framework, we optimize the charge and discharge operations for both BES and HES.

\subsection{BES}
For the optimal scheduling of BES, the requisite dynamic model for the charge and discharge operations is defined as follows,
\begin{equation}
	\begin{aligned}\label{1}
		E_{t+1}^{\rm SoC} = E_t^{\rm SoC} + \eta^{\rm Bat} P_t^{\rm Bat} \Delta t,
	\end{aligned}
\end{equation}
where $E_t^{\rm SoC}$ represents the current SoC of the battery. $\eta^{Bat}$ signifies the charge/discharge efficiency, where $\eta^{Bat}=0.9$ for charging and $\eta^{Bat}=0.95$ for discharging. $P_t^{\rm Bat}$ denotes the charge/discharge power of the BES equipment, with $P_t^{\rm Bat}>0$ indicating charging and $P_t^{\rm Bat}<0$ indicating discharging at time $t$. It's important to note that we use a time interval of $\Delta t = 1$ hour, during which all values are assumed to remain constant.

In accordance with the dynamic model described above, the BES must adhere to the following operational constraints,
\begin{subequations}
	\begin{align}\label{2}
	E_{\rm min}^{\rm SoC} \leq &E_t^{\rm SoC}\leq E_{\rm max}^{\rm SoC}, \tag{2a}\\
	P_{\rm min}^{\rm Bat} \leq &P_t^{\rm Bat}\leq P_{\rm max}^{\rm Bat}, \tag{2b}
    \end{align}
\end{subequations}
where $E_{\rm min}^{\rm SoC}$ and $E_{\rm max}^{\rm SoC}$ represent the minimum and maximum energy states of the battery. $P_{\rm min}^{\rm Bat}$ and $P_{\rm max}^{\rm Bat}$ are the limits of the charge/discharge power per unit time.

To maximize the benefits of the PV-ESS operator, it is crucial to have an accurate and easily solvable battery cost function that accounts for coupled capital, degradation, and operation costs. Both degradation and operation costs are intertwined with the capital cost. Over time, the battery experiences degradation from its original state, with the degradation rate being dependent on operating characteristics and conditions. Thus, we employ degradation cost to encompass both capital and operation costs. As described in prior studies \cite{10039495, 8805438}, the degradation cost incurred during operation is significantly influenced by various factors, including battery capacity, SoC limits, environmental temperature, and current. In our BES cost function, we assume the presence of a temperature control system in the environment, thereby neglecting the impact of high temperatures. We also establish appropriate minimum and maximum values for SoC while not considering the effects of SoC limits. Instead, we focus on the degradation associated with DoD and discharge rate during periodic charge and discharge processes within the electricity market framework.

It is essential to highlight that the impact of DoD on degradation costs is influenced not only by the difference SoC at adjacent times but also by the initial and final levels of SoC during discharge process. For instance, discharging from 70$\%$ to 0$\%$ results in more significant degradation compared to discharging from 100$\%$ to 30$\%$. Consequently, we employ SoC instead of DoD in the BES cost function.

Additionally, the discharge rate $v^{\rm DCR}_t$ related to the current can be computed as follows,
\begin{equation}
	v^{\rm DCR}_t=\frac{E_{t-1}^{\rm SoC}-E_t^{\rm SoC}}{\Delta t}.
\end{equation}

Building on insights from \cite{8805438} and taking into consideration the above mentioned influencing factors, the BES cost $C^{\rm Bat}$ is expressed as follows,
\begin{equation}\label{cost}
	C^{\rm Bat}_t=\frac{c^{\rm Bat}_{\rm cc}}{{\rm Cap}\eta^2_{\rm r}\phi}((1-E_t^{\rm SoC})^{\omega}-(1-E_{t-1}^{\rm SoC})^{\omega}),
\end{equation}
where $c^{\rm Bat}_{\rm cc}$ represents the BES capital cost of the battery. $\rm Cap$ means the battery capacity, and $\eta_{\rm r}$ signifies the round trip efficiency of the BES. Coefficients $\phi$ and $\omega$ are used to capture the relationship between DoD and the number of cycles. 

To simplify the BES cost function and make it readily integrable into a comprehensive cost function of the heterogeneous PV-ESS, the cost function from Eq. (\ref{cost}) can be linearized as follows,
\begin{equation}
	C^{\rm Bat}_t=w_1 E_t^{\rm SoC} + w_2 E_{t-1}^{\rm SoC} + w_3 v^{\rm DCR}_t + w_4,
\end{equation}
where $w_1$ and $w_2$ are the coefficients of the cost related to DoD. $w_3$ is the coefficient of the cost related to discharge rate. $w_4$ is related to battery capacity and serves as a linearization offset term within the degradation cost function \cite{8805438}. Indeed, it's important to emphasize that the proposed degradation cost function for the BES is time-dependent, as it takes into account factors such as DoD, SoC, and discharge rate, all of which vary with time. These parameters collectively enable a comprehensive and time-sensitive representation of the degradation cost model.

\subsection{HES}
In the pursuit of optimizing the scheduling of the HES, we first introduce the composition of the HES and elucidate the energy conversion processes associated with each component. Subsequently, the dynamic model of the HES that govern these energy conversion processes is presented. Finally, we delve into the HES cost function.

The HES comprises hydrogen proton exchange membrane fuel cell stacks (FCs), an electrolyzer (EL), and a hydrogen storage reservoir, encompassing the conversion of hydrogen into electricity and vice versa \cite{9857842}. More specifically, the hydrogen FCs serve as power generation equipment capable of converting the chemical energy stored in hydrogen into electric energy, while the EL can perform the reverse operation. We quantify these conversion relationships using the molar flow of hydrogen and the power output of the EL and FCs, in accordance with Faraday's law \cite{cau2014energy},
\begin{subequations}
	\begin{align}
		& F_t^{\mathrm{EL}_{\rm H_2}}=\eta^{\mathrm{EL}} P_t^{\mathrm{EL}} / {\rm NCV_{\rm H_2}}, \tag{6a}\\
		& F_t^{\mathrm{FC}_{\rm H_2}}=P_t^{\mathrm{FC}} / \eta^{\mathrm{FC}} {\rm NCV_{\rm H_2}}, \tag{6b}
    \end{align}
\end{subequations}
where $F_t^{\mathrm{EL}_{\rm H_2}}$ and $F_t^{\mathrm{FC}_{\rm H_2}}$ represent the hydrogen molar flow in the EL and FCs, respectively. $P_t^{\mathrm{EL}}$ and $P_t^{\mathrm{FC}}$ denote the power output of the EL and FCs, while $\eta^{\mathrm{EL}}$ and $\eta^{\mathrm{FC}}$ characterize the energy conversion rates of the EL and FCs, respectively. ${\rm NCV_{\rm H_2}}$ represents the net calorific value, which is the effective calorific value obtained by subtracting the heat of water vaporization from the full combustion calorific value.

Based on the above conversion relationship, the state of the hydrogen storage reservoir at the previous time step and the change in the hydrogen molar flow at current time step can be employed to calculate the current state of the hydrogen storage reservoir in the following manner \cite{9857842},
\begin{equation}
	\begin{aligned}\label{3}
		E_t^{\rm LoH}= & \left(1-\eta^{\mathrm{HES}}\right) E_{t-1}^{\rm LoH} +\frac{\mathcal{R} T_{\rm H_2}}{V_{\rm H_2}}\left(F_t^{\mathrm{EL}_{\rm H_2}}-F_t^{\mathrm{FC}_{\rm H_2}}\right),
	\end{aligned}
\end{equation}
where $E_t^{\rm LoH}$ denotes the pressure of the hydrogen storage reservoir at time $t$. $\eta^{\mathrm{HES}}$ represents the self-consumption rate of the hydrogen storage equipment. $\mathcal{R}$, $T_{\rm H_2}$, and $V_{\rm H_2}$ correspond to the gas constant, mean temperature of the hydrogen storage reservoir, and the reservoir volume, respectively. Similar to the BES, we still assume the existence of a temperature control system, so the mean temperature remains constant.

Additionally, the HES must adhere to the following operational constraints,
\begin{subequations}
	\begin{align}
		& E_{\rm min}^{\rm LoH} \leq E_t^{\rm LoH} \leq E_{\rm max}^{\rm LoH}, \tag{8a}\\
		& P_{\rm min}^{\mathrm{EL}} \leq P_t^{\mathrm{EL}} \leq P_{\rm max}^{\mathrm{EL}}, \tag{8b}\\
		& P_{\rm min}^{\mathrm{FC}} \leq P_t^{\mathrm{FC}} \leq P_{\rm max}^{\mathrm{FC}}, \tag{8c}\\
		&\quad P_t^{\mathrm{EL}}P_t^{\mathrm{FC}} =0, \tag{8d}
	\end{align}
\end{subequations}
where $E_{\rm min}^{\rm LoH}$ and $E_{\rm max}^{\rm LoH}$ are the lower and upper limits for the pressure of the hydrogen storage reservoir. $P_{\rm min}^{\mathrm{EL}}$ and $P_{\rm max}^{\mathrm{EL}}$ impose constraints on the power output of the EL, while $P_{\rm min}^{\mathrm{FC}}$ and $P_{\rm max}^{\mathrm{FC}}$ are limits for the power output of FCs. The final constraint specifies that the EL and FCs cannot operate simultaneously at time $t$.

Following the depiction of the state transition of the hydrogen storage reservoir, we now turn our attention to the comprehensive cost function of the HES. Much like the BES, the cost of HES incorporates capital, degradation, and operation costs. In contrast to the BES, HES incurs a higher capital cost. It's evident that the latter two costs are intricately tied to the operational scheduling of the HES, which includes factors like runtime, state switching frequency, power output, and current. Inspired by \cite{9399252}, the cost function for the HES can be formulated as follows,
\begin{equation}
	\begin{aligned}
		C^{\rm HES}_t= \sum_{i={\rm EL, FC}}\left(\left(\frac{c_{\rm cc}^i}{\nu^i}+c_{\rm op}^i\right) {\sigma}^i +c_{\rm st}^i {\zeta}^i + c_{\rm de}^i {\kappa}^i \right),
	\end{aligned}
\end{equation}
where ${\sigma}^i \in \{{\sigma}^{\rm EL}$, ${\sigma}^{\rm FC}\}$ are binary variables associated with the on/off-status of EL and FCs, where 0 indicates off-status and 1 indicates on-status. ${\zeta}^i$ represent logical variables that account for the start-up state. ${\kappa}^i$ is defined as the power variation at instances when EL/FCs are active. $c_{\rm cc}^i$ and $\nu^i$ denote the capital acquisition cost for the EL/FCs devices and the total number of working hours. $c_{\rm op}^i$ is the hourly operation cost associated with the maintenance of EL/FCs devices. $c_{\rm st}^i$ and $c_{\rm de}^i$ are utilized to formulate the degradation cost resulting from start-up cycles and high current values during the charge/discharge processes. 

\subsection{MDP Formulation}
As for the heterogeneous PV-ESS scheduling framework developed in this paper shown in Fig. \ref{Framework of energy market with heterogeneous ESS}, it comprises an operator and a heterogeneous ESS integrated with PV, which can serve as both an energy supplier and an energy consumer in the energy market. On the supply side, the framework primarily includes the traditional centralized main grid, which relies on thermal power generation. The market electricity prices in this framework are determined by the main grid. On the demand side, there are various users with diverse power requirements, such as municipalities, factories, and individual households. We assume a continuous electricity demand scenario, ensuring that users are constantly in need of electricity from a whole perspective.

It's crucial to emphasize that, to enhance the competitiveness of the PV-ESS in the energy market, its transaction prices consistently remain below market prices. This allows users prioritize purchasing electricity at a lower price from the PV-ESS. The revenue of the PV-ESS operator is derived from the sale of PV power and energy arbitrage. Energy arbitrage entails storing excess PV power or procuring electricity from the main grid when market prices are low and subsequently selling it at a lower price than the market rate when prices rise and electricity demand is high. Typically, the selling price is often higher than the price at which the PV-ESS initially bought electricity from the market.

The operator has access to energy market information, including electricity prices, as well as internal status information about the PV-ESS. This internal status information covers PV power generation, the SoC of the BES, the hydrogen storage level of the HES, and the operational status of each equipment. This framework forms the basis for optimizing operations of the heterogeneous PV-ESS and maximizing its economic profitability.

To design an explainable scheduling strategy for the heterogeneous PV-ESS, the charge and discharge operation scheduling problem can be formulated as a MDP.  In this formulation, state transitions depend solely on the previous one step state and not on any memory. The MDP framework comprises four key elements: a set of states ($s\in \mathbb{S}$), a set of actions ($a\in \mathbb{A}$), a reward function $r$, and transition probabilities $p$ from state $s$ and action $a$ to state $s'$ \cite{sutton2018reinforcement}. For the operation scheduling problem, these elements are defined as follows:
\subsubsection{The state}
The state $s_t$ serves as a representation of the current situation of the heterogeneous PV-ESS. In this study, the state encompasses the following elements,
\begin{equation}
	\begin{aligned}
		s_t = \{{\rm Pr}_t, P_t^{\rm PV}, E_t^{\rm SoC}, E_t^{\rm LoH}, {\sigma}^{\rm EL}, {\sigma}^{\rm FC}\},
	\end{aligned}
\end{equation}
where ${\rm Pr}_t$ represents the dynamic electricity price, and the $P_t^{\rm PV}$ signifies the power output from PV generation. In order to ensure that the EL and FCs do not operate simultaneously, we impose the constraint ${\sigma}^{\rm EL} {\sigma}^{\rm FC} = 0$. The observation is denoted as $\{{\rm Pr}_t, P_t^{\rm PV}, E_t^{\rm SoC}, E_t^{\rm LoH}\}$.

\begin{figure*}[htb]	
	\centering	
	\includegraphics[scale=1, width=7in]{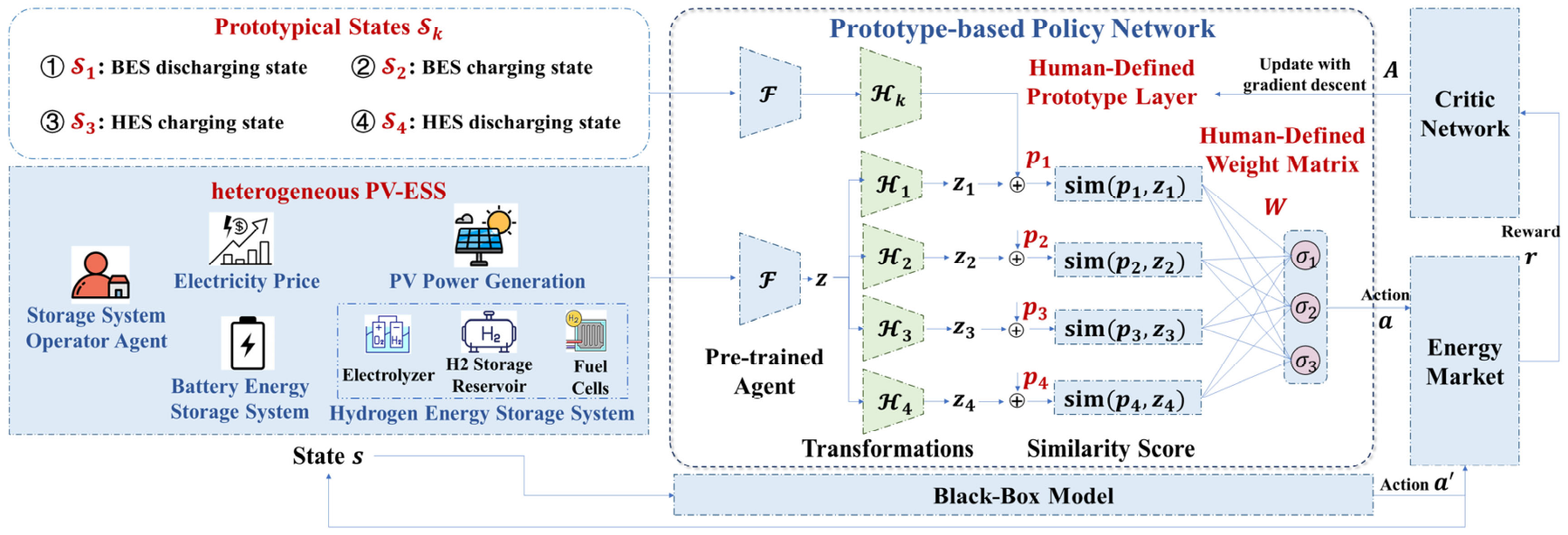}	
	\caption{Structure of interpretable DRL method with a prototype-based policy network.}	
	\label{framework}
\end{figure*}

\subsubsection{The action}
Based on the definition of the system state, the actions are defined as follows,
\begin{equation}
	\begin{aligned}
		a_t=\{P_t^{\rm Bat}, P_t^{\rm EL}, P_t^{\rm FC}\},
	\end{aligned}
\end{equation}
where $P_t^{\rm Bat}$, $P_t^{\rm EL}$, and $P_t^{\rm FC}$ are continuous variables in the action space $\mathbb{A}$. It is important to take into account the constraints imposed by the battery capacity, hydrogen storage reservoir pressure, the charge/discharge power limitations, and the power output of the EL/FCs. Consequently, the actual actions are constrained as follows,
\begin{subequations}
	\begin{align}
		& P_t^{\rm Bat} = \begin{cases} {\rm min}\{P_t^{\rm Bat}, \frac{1-E_t^{\rm SoC}}{\eta^{\rm Bat} \Delta t}\}, & \text {if}\quad P_t^{\rm Bat}>0, \\ {\rm max}\{P_t^{\rm Bat}, \frac{-E_t^{\rm SoC}}{\eta^{\rm Bat} \Delta t}\}, & \text {if}\quad P_t^{\rm Bat}<0,\end{cases} \tag{12a}\\
		& \begin{cases}P_t^{\rm EL} = {\rm min}\{P_t^{\rm EL}, \frac{\Delta E_t^{\rm LoH}V_{\rm H_2}}{\mathcal{R} T_{\rm H_2}}\}, & \text {if}\quad P_t^{\rm EL}>0, \\ 
		P_t^{\rm FC} = {\rm min}\{P_t^{\rm FC}, \frac{\Delta E_t^{\rm LoH} V_{\rm H_2}}{\mathcal{R} T_{\rm H_2}}\}, & \text {if}\quad P_t^{\rm FC}>0, \end{cases} \tag{12b}\\
	    & \Delta E_t^{\rm LoH} = \begin{cases} E_{\rm max}^{\rm LoH} - \left(1-\eta^{\mathrm{HES}}\right)E_t^{\rm LoH}, & \text {if}\quad P_t^{\rm EL}>0, \\ \left(1-\eta^{\mathrm{HES}}\right)E_t^{\rm LoH}, & \text {if}\quad P_t^{\rm FC}>0. \end{cases} \tag{12c}
	\end{align}
\end{subequations}
Eq. (12a) serves to ensure that the charge and discharge power of the battery do not breach the maximum/minimum capacity limits. Additionally, Eq. (12b) ensures that the hydrogen produced by electrolysis does not exceed the maximum remaining capacity of the hydrogen storage tank, and it also ensures that the hydrogen demand of fuel cell stacks does not exceed the available hydrogen reserve. Eq. (12c) calculates the permissible pressure state change while taking into account the impact of equipment self-consumption.

\subsubsection{State transition}
The system transition at time $t$ can be depicted as Eq. (\ref{1}) and Eq. (\ref{3}).
	
\subsubsection{The reward function}
The reward function is designed to quantify benefits of the PV-ESS operator at time $t$, aligning with the optimization objective,
\begin{subequations}
	\begin{align}
		r_t=&\rho {\rm Pr}_t P_t^{\rm sell}-C^{\rm Bat}_t-C^{\rm HES}_t, \tag{13a} \\
		P_t^{\rm sell}=&P_t^{\rm PV} + P_t^{\rm FC} - P_t^{\rm Bat} - P_t^{\rm EL}, \tag{13b}
	\end{align}
\end{subequations}
where $P_t^{\rm sell}$ represents the electricity sold to customers. $\rho \in (0, 1]$ signifies the discount rate applied to the market electricity price ${\rm Pr}_t$. For this analysis, $\rho$ is set to 0.95. This choice indicates that it is more advantageous for customers to engage in energy transactions with the PV-ESS operator rather than with the power grid, primarily due to the more favorable electricity prices offered by the PV-ESS operator. Participation in the energy market with the PV-ESS operator clearly leads to improved economic performance for prosumers.

\section{Proposed approach}
In this section, we provide a detailed introduction to the proposed interpretable DRL method, which includes a prototype-based policy network designed for pre-hoc interpretability. We will first delve into the prototype-based policy network, and then offer insights into the human-friendly interpretable DRL method, which demonstrates how the prototype-based policy network enhances the transparency and understandability of the agent's decision-making process.

\subsection{Prototype-based Policy Network}
The rapid advancement of DRL across various domains has led to the emergence of interpretable methods to facilitate its real-world application. Currently, the prevalent methods are post-hoc interpretation techniques that provide insights into model predictions over time \cite{10068282}. While these methods are widely adopted, they may not provide a complete understanding of the agent's decision-making process, as it remains concealed.

Motivated by this, we introduce a prototype-based policy network that transforms a DRL agent from a black-box model into an interpretable model \cite{kenny2022towards}. This approach compels the agent to generate policies that are comprehensible in a human-friendly manner. The structure of the prototype-based policy network, as applied to the PV-ESS scheduling problem, is depicted in Fig. \ref{framework}. It comprises a pre-trained agent serving as a coding network, several transformation networks along with their corresponding prototypical states. The similarity score is derived by comparing the prototypes transformed from prototypical states with the actual potential representation of the states. This score is then employed to guide the agent's decision-making process. Notably, the method's interpretability is derived from prototypes based on human experience, which incorporate intuitive and easily understandable actions in the prototypical states. These prototypes, in turn, provide guidance for the actual actions within each dimension.

\begin{remark}
It's important to note that a pre-trained agent can be acquired using a black-box approach, which can achieve commendable performance. The primary purpose of the prototype-based policy network is to assist the pre-trained agent in rendering its decision-making process transparent and understandable, thereby enhancing the pre-hoc interpretability of the algorithm. Consequently, within the prototype-based policy network, we fully leverage the capabilities of the pre-trained agent.
\end{remark}

We define the policy derived from the pre-trained agent, based on the black-box model, as $\pi'$ and assume that this policy can be decomposed into an encoder network $\mathcal{F}$ and a linear layer, implying that $\pi'=W'\mathcal{F}(s)+b'$ \cite{kenny2022towards}. To fully utilize the well-performing black-box model, within the prototype-based policy network, we initially input the state $s$ into the pre-trained encoder network $\mathcal{F}$ to obtain the latent representation $z=\mathcal{F}(s)$. Subsequently, to elucidate the action generation process clearly, separate transformation networks $\mathcal{H}_k$ are introduced for each action dimension, which map the latent representation $z$ of the state $s$ to specific representations $z_k$ for different action dimension $k$. In particular, for the PV-ESS scheduling problem, the action encompasses three dimensions: the charge/discharge power of the BES, the power output of the EL, and the power output of the FCs. However, to distinguish between the charge and discharge behavior of the BES, the first dimensional action $P^{\rm Bat}$ is divided into separate components for charging and discharging, resulting in a total of four action dimensions. This division enhances the ease of prototype design and facilitates a better understanding of the agent's decision-making process. $$z_k = \mathcal{H}_k(z), k \in \{1, 2, 3, 4\}.$$ 

\begin{figure}[htb]	
	\centering	
	\includegraphics[scale=1, width=3.5in]{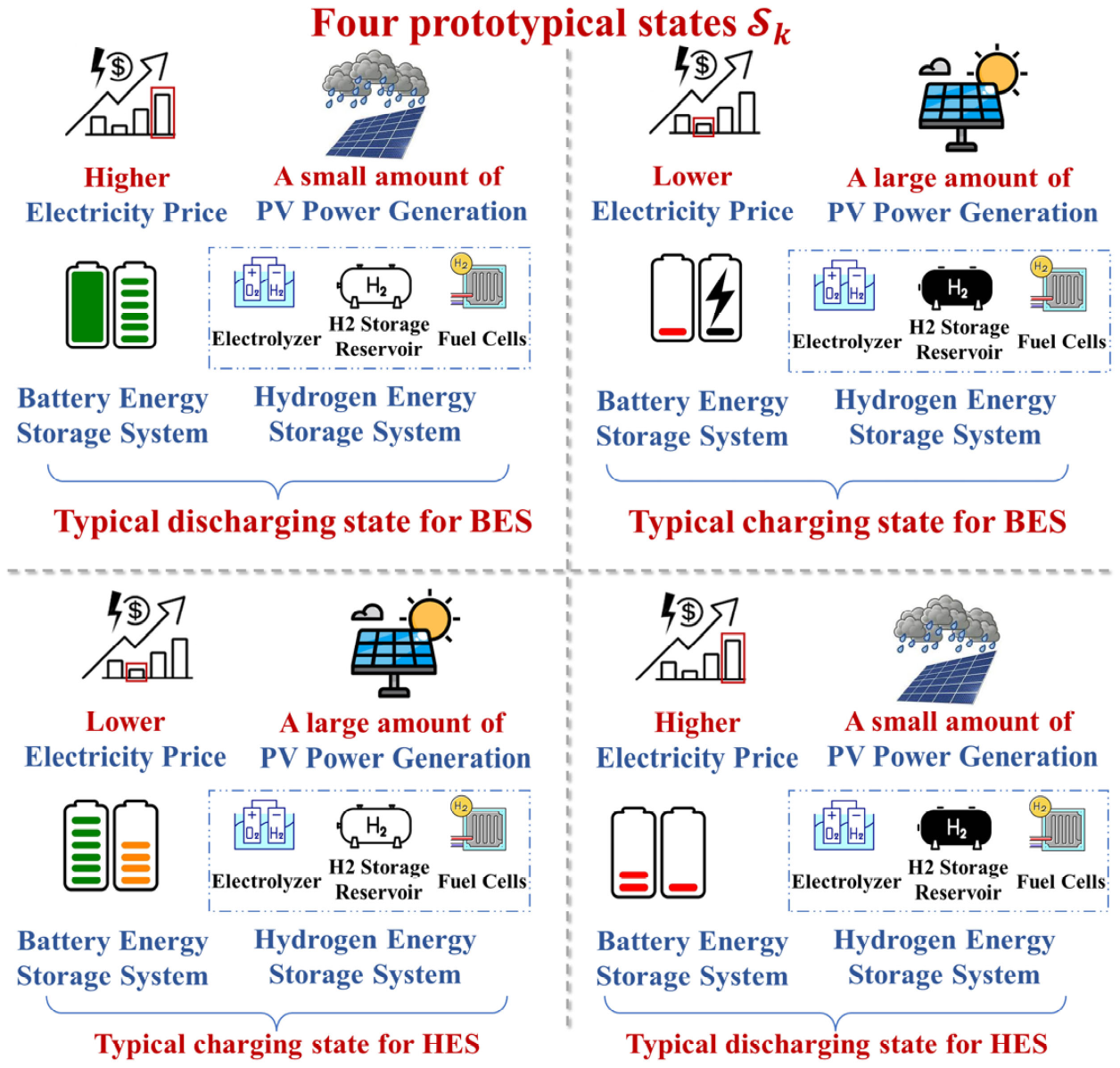}	
	\caption{Four prototypical states.}	
	\label{prototype}
\end{figure}

With the networks described above, a set of prototypical states $\mathcal{S}_k$ are designed for prototypical actions within each dimension, which are intuitive and human-friendly. These prototypical states are then fed into both the original encoder network $\mathcal{F}$ and the transformation networks $\mathcal{H}_k$, and used as prototypes $p_k=\mathcal{H}_k(\mathcal{F}(\mathcal{S}_k))$ for the $k$-th dimension. For the PV-ESS scheduling problem, we design four prototypical states, as illustrated in Fig. \ref{prototype}. One prototypical state represents a typical charging scenario for the BES in an environment characterized by low electricity prices, ample PV power generation, and a low level of battery energy. Conversely, another prototypical state signifies an obvious and intuitive profitable operation for the BES, which is discharging in an environment featuring high market electricity prices, insufficient PV power generation, and a high SoC. Similar situations apply to the EL and FCs within the HES as well.

Utilizing the prototypes $p_k$ mentioned earlier, the similarity between specific representations $z_k$ and prototypes $p_k$ are calculated as outlined in \cite{chen2019looks}. Subsequently, we introduce a human-defined linear weight matrix $W$ that is employed in combination with the similarity scores to generate actions. This weight matrix $W$ encapsulates the relationship between prototypes and actions, and it provides an intuitive explanation for $W_k$, which signifies how the prototype $p_k$ should influence the action $a_k$. This approach ensures that each prototype is associated with an action that is intuitively comprehensible.
\begin{subequations}
	\begin{align}
		{\rm sim}(z_k,p_k)=&{\rm log}\Bigg(\frac{(z_k-p_k)^2+1}{(z_k-p_k)^2+\epsilon}\Bigg), \tag{13a} \\
		a_{t,k}=&W_k{\rm sim}(z_k,p_k), \tag{13b}
	\end{align}
\end{subequations}
where $\epsilon=1e^{-5}$ is a hyperparameter utilized to characterize similarity.

In the scheduling problem of the PV-ESS, where $P_t^{\rm Bat}$ is the charge/discharge power, we set $W_1 = 1$ and $W_2 = -1$. This configuration signifies that the charge/discharge power should be equal to the difference between the similarity to prototypical charge and discharge actions. Here, we illustrate with a straightforward example that when the current state of the BES closely resembles the typical BES charging state as shown in Fig. \ref{prototype}, the action $P_t^{\rm Bat}$ becomes strongly associated with charging. Likewise, if the actual state bears similarity to the typical BES discharge state, the learned action leans towards discharge. Consequently, the decision-making process becomes more interpretable as it naturally explains why a specific action is chosen. This approach can be likened to a case-based reasoning strategy, where the decision to take action $a$ is made because the current situation bears similarity to a prior prototypical situation in which action $a$ was also chosen \cite{kenny2022towards}. $$P_t^{\rm Bat} = W_1{\rm sim}(z_1,p_1)+W_2{\rm sim}(z_2,p_2).$$

\begin{remark}
	It's important to emphasize that, in contrast to previous approaches where prototypes are learned \cite{chen2019looks}, the prototypes in our method are human-defined. This choice is in line with the idea that involving humans in the learning loop can be beneficial, as suggested by \cite{bontempelli2022concept}. Similarly, the weight matrix $W$ is manually defined rather than learned. Learning $W$ could lead to each prototype corresponding to multiple undesirable actions, which is why we opt for manual specification.
\end{remark}

In the prototype-based policy network, only the transformation networks $\mathcal{H}_k$ with parameters $\psi$ can be trained. These networks build upon the pre-trained encoder network $\mathcal{F}$, the provided prototypical states $\mathcal{S}_k$, and a manually specified weight matrix $W$. The training process involves minimizing the loss between the output of the prototype-based policy network $a$ and the action $a' \in \pi'$ obtained from the black-box model in specific states. The parameters $\psi$ of the transformation networks $\mathcal{H}_k$ are updated using the gradient descent method. The pseudocode for training the prototype-based policy network is presented in the following Algorithm \ref{algorithm}.

\begin{algorithm}[htb]  
	\caption{Training the prototype-based policy network}  
	\label{algorithm}  
	\begin{algorithmic}[1]  
		\Require
		A pre-trained agent with encoder network $\mathcal{F}$ and policy $\pi'$.
		\Ensure  
		A well-trained prototype-based policy network.
		\State Initialize the prototype-based policy network with a manually specified weight-matrix $W$.
		\State Sample $n$ state-action pairs from Dataset collected by the pre-trained agent $\mathcal{D}\leftarrow \{(s, \pi'(s))\}_{j=0}^n$.
		\State Choose Human-Interpretable Prototypical States $\mathcal{S}_k \in \mathbb{S}$.
		\For{batch $(s,a') \in \mathcal{D}$}
		\State $z= \mathcal{F}(s)$;
		\For{$k \in \{1, 2, 3, 4\}$}
		\State $a_k = 0$;
		\For{each $\mathcal{S}_k$}
		\State $p_k=\mathcal{H}_k(\mathcal{F}(\mathcal{S}_k)) $
		\EndFor
		\State $z_k=\mathcal{H}_k(z)$
		\State $a_k=a_k+W_k{\rm sim}(z_k,p_k)$
		\State Minimize Loss $\mathcal{L}(a | a', \mathcal{F}, \psi, W)$ with gradient descent, updating only $\psi$.
		\EndFor
		\State Cache all $p_k=\mathcal{H}_k(\mathcal{F}(\mathcal{S}_k))$ for testing time inference.
		\EndFor
		\State \textbf{return} trained prototype-based policy network.
	\end{algorithmic}  
\end{algorithm}

\subsection{Interpretable DRL Algorithm}
Before training a prototype-based policy network, a well-trained black-box model is required. This black-box model is used, in part, as an encoder network $\mathcal{F}$ for the interpretable policy network. The PPO algorithm is employed for pre-training the agent. In this section, we will provide a brief introduction to the PPO algorithm as applied to solve the scheduling problem of the PV-ESS.

The PPO algorithm employs a neural network architecture with shared parameters $\theta$ for predicting the policy function and the value function. The loss function used to train the shared network encompasses error terms from the policy surrogate and the value function. Additionally, an entropy term is incorporated into the loss function to promote exploration in the action space. The loss function can be expressed as follows,
\begin{equation}
	\begin{aligned}\label{loss}
		L_t(\theta)=\hat{\mathbb{E}}_t\left[L_t^{\rm C}(\theta)-m_1 L_t^{\rm V}(\theta)+m_2 L_t^{\rm S}\left[\pi_\theta|\left(s_t\right)\right]\right],\\
	\end{aligned}
\end{equation}
where $L_t^{\rm C}(\theta)$ represents the policy surrogate error term, $L_t^{\rm V}(\theta)=\left(V_\theta\left(s_t\right)-V_t^{\operatorname{targ}}\right)^2$ is the error of the value function, and $L_t^{\rm S}\left[\pi_\theta|\left(s_t\right)\right]$ is the entropy bonus used for exploration. $m_1$ and $m_2$ are coefficients for $L_t^{\rm V}$ and $L_t^{\rm S}$, respectively. $L_t^{\rm C}(\theta)$ can be calculated as follows,
\begin{equation}
	\begin{aligned}
		L^{\rm C}_t(\theta)=\hat{\mathbb{E}}_t\left[\min \left(\Upsilon_t \hat{A}_t, \operatorname{clip}\left(\Upsilon_t, 1- \xi, 1+ \xi \right) \hat{A}_t\right)\right],
	\end{aligned}
\end{equation}
where $\hat{A}_t=\delta_t+(\gamma \lambda) \delta_{t+1}+\cdots+\cdots+(\gamma \lambda)^{T-t+1} \delta_{T-1}$ is the advantage function estimated with $\delta_t=r_t+\gamma V\left(s_{t+1}\right)-V\left(s_t\right)$. $\Upsilon_t=\frac{\pi_\theta(a_t|s_t)}{\pi_{\theta_{\rm old}}(a_t|s_t)}$ is the probability ratio, and $\xi$ is a hyperparameter.

With the loss function as Eq. (\ref{loss}), the shared network with parameters $\theta$ for both policy and value functions can be trained as shown in \cite{schulman2017proximal}.

\section{Experiments}
This section outlines the experiments to assess the effectiveness of the proposed interpretable DRL method for solving the heterogeneous PV-ESS scheduling problem. Meanwhile, the revenue of the PV-ESS operator in various cases is also analyzed, considering heterogeneous energy storage devices. The following subsections describe the preliminaries of experiments. Then, the performance of the proposed interpretable DRL method compared with other approaches is verified and corresponding discussions about the pre-hoc interpretability are provided. Lastly, we examine the revenue of the PV-ESS operator with heterogeneous energy storage devices in various scenarios and assess the impact of the learning rate on convergence and optimality.

\subsection{Preliminaries of Case Studies}
The experiments are conducted using PV power generation data and time-varying electricity prices data from \cite{data123}. We focus on the PV-ESS scheduling problem during a single day, with each hour representing one time slot, resulting in the time horizon $T=24$ hours. Considering the one-hour time scale, there are a total of 8760 data sets in a year. As for the heterogeneous PV-ESS, the initial SoC is set to a random value between 0.25 and 1, while the initial level of hydrogen storage (LoH) is set to a random value between 5 and 35. Various parameters used in the experiments are detailed in Table \ref{table1}.

We design four distinct cases, as summarized in Table \ref{table2}, to analyze the impact of the heterogeneous PV-ESS on the operator's revenue, while considering various characteristics and cost structures. The experiments are conducted using Python 3.6.13 with the machine learning library PyTorch 1.8.1.

\begin{table}[thp]\footnotesize
	\centering
	\caption{Parameters Used in the Experiments.} \label{table1}
	\addtolength{\tabcolsep}{4.8pt}
	\begin{tabular}{cc|cc}
		\toprule
		 Parameters     &    Value  &  Parameters     &    Value      \\
		\midrule
		$a$     &    -36.23   & 	${m_1}$     &   0.5      \\
		$b$     &    34.80   & 	${m_2}$     &    0.01   \\
		$c$     &    2.77      &  	$\eta^{\rm EL}$     &    0.725            \\
		$d$     &   -2.45      &	$\eta^{\rm FC}$     &   0.6           \\
	 	$\xi$   & 0.2       &     $\eta^{\rm HES}$    &    0.05         \\
	    $\lambda$  & 0.95   & 	 ${V_{\rm H_2}}$     &    35 Nm$^3$    \\
		$\gamma$  & 0.99 & ${T_{\rm H_2}}$     &    313 K   \\
		$\mathcal{R}$ &  8.314 J/mol K   & 	 ${\rm NCV_{\rm H_2}}$     &    240 MJ/kmol        \\
		\bottomrule
	\end{tabular}
\end{table}

\begin{table}[thp]\footnotesize
	\centering
	\caption{Description of Experimental Cases.} \label{table2}
	\addtolength{\tabcolsep}{4.8pt}
	\begin{tabular}{c|cc|cc}
		\toprule
		\multirow{2}{*}{Cases}   & \multicolumn{2}{c|}{BES}  & \multicolumn{2}{c}{HES} \\
		   &      existence     &    cost      &      existence     &     cost     \\
		\midrule
		Case 1     &    \checkmark       &       \checkmark   &   \checkmark        &   \checkmark  \\
	Case 2    &     \checkmark      &     $\times$     &      \checkmark     &  $\times$   \\
		Case 3     &     \checkmark      &       \checkmark   &    $\times$       &      $\times$       \\
		Case 4   &     $\times$      &      $\times$    &       \checkmark    &     \checkmark      \\
		\bottomrule
	\end{tabular}
\end{table}

\subsection{Performance of the Prototype-based Policy Network}
Here, we first present several baseline methods employed for comparison with the proposed interpretable DRL method using the prototype-based policy network in the scheduling problem of the heterogeneous PV-ESS, following the presentation of four human-designed prototypes. Then the performance and interpretability of each method are presented, and the results are analyzed to provide insights into the benefits of the proposed interpretable DRL approach. It is remarkable that the pre-trained agent is based on PPO, and only the design of the policy network and prototypes is changed in different baselines. 

Below, we provide an overview of the baseline methods:
\begin{itemize}
	\item Prototype-based policy network*: This variant uses a single transformation network $\mathcal{H}$ for all prototypes, rather than individual transformation networks for each prototype dimension. Besides, it learns prototypes as training parameters, which are then mapped to the most recent training example. In contrast, our proposed prototype-based policy network employs manually defined prototypes. The purpose of this variant is also to conduct an ablation test, exploring the impact of these specific design choices, such as separate transformation networks and manually defined prototypes, on the performance and interpretability of the prototype-based policy network.
\end{itemize}

\begin{itemize}
	\item K-Means: This method obtains the prototypes through the clustering and the mapping process. The clustering process aims to identify centroids that match the number of prototypes used in the proposed prototype-based policy network. When clusters in space $z$ with the same number as the prototypes are obtained, each centroid is mapped to the most recent training sample, essentially associating each centroid with a specific state from the training data. These states serve as the prototypical states. Besides, K-Means are allowed to learn the weight parameters of the last layer, which suggests that K-Means clusters are not only used to identify prototypes but also contribute to the network's final decision-making process through weight parameters. 
\end{itemize}

\begin{table*}[thp]\footnotesize
	\centering
	\caption{Results of the Prototype-based Policy Network compared with Various baselines.} \label{table3}
	\addtolength{\tabcolsep}{4.8pt}
	\begin{threeparttable}
	\begin{tabular}{cc|cccc}
		\toprule
		Methods & Metrics   & Case 1 & Case 2 & Case 3 & Case 4 \\ \midrule
		\multirow{2}{*}{\begin{tabular}[c]{@{}l@{}}Prototype-based \\  Policy Network\end{tabular}}  & Reward & \textbf{ 7.43$ \pm $ 3.25 }&\textbf{ 23.63  $ \pm $ 1.01}  & \textbf{15.69 $ \pm $0.78 } & \cellcolor{lightgray} 7.48 $ \pm $ 0.67  \\
		& MSE & \textbf{ 4.92 $ \pm $ 1.80} & \textbf{ 4.97 $ \pm $ 0.51} & \textbf{ 3.17 $ \pm $ 0.50 } &  \textbf{0.13 $ \pm $0.07}  \\ \cline{1-2}
		\multirow{2}{*}{\begin{tabular}[c]{@{}l@{}}Prototype-based\\  Policy Network*\end{tabular}} & Reward & -7737.81  $ \pm $ 1627.94 & -6979.15 $ \pm $ 4963.04  & -5054.70 $ \pm $  1746.51 & -5943.58 $ \pm $  3174.58 \\
		& MSE & 58.26$ \pm $6.49  & 45.46 $ \pm $8.92 & 138.61 $ \pm $37.65 & 10.60$ \pm $2.92 \\ \cline{1-2}
		\multirow{2}{*}{K-Means} & Reward & \cellcolor{lightgray}-98.36$ \pm $ 67.68 & \cellcolor{lightgray}14.72 $ \pm $ 3.46 & \cellcolor{lightgray}13.94 $ \pm $  0.72 & \textbf{ 9.38 $ \pm $  3.49  } \\
		& MSE & \cellcolor{lightgray}24.74 $ \pm $  3.99 & \cellcolor{lightgray}25.73 $ \pm $ 3.12 & \cellcolor{lightgray}23.37 $ \pm $  1.25 & \cellcolor{lightgray}8.44 $ \pm $   2.55 \\ \cline{1-2}
	\multicolumn{2}{c|}{Reward of Black-box}&  12.42 & 24.44 &  14.70 & 7.66 \\ 
		\bottomrule
	\end{tabular}
 \begin{tablenotes}
	\footnotesize
	 \item Results in bold and cells colored gray denote the best and the second best, respectively.
\end{tablenotes}
\end{threeparttable}
\end{table*}

\begin{figure}[htb]	
	\centering	
	\includegraphics[scale=1, width=3.5in]{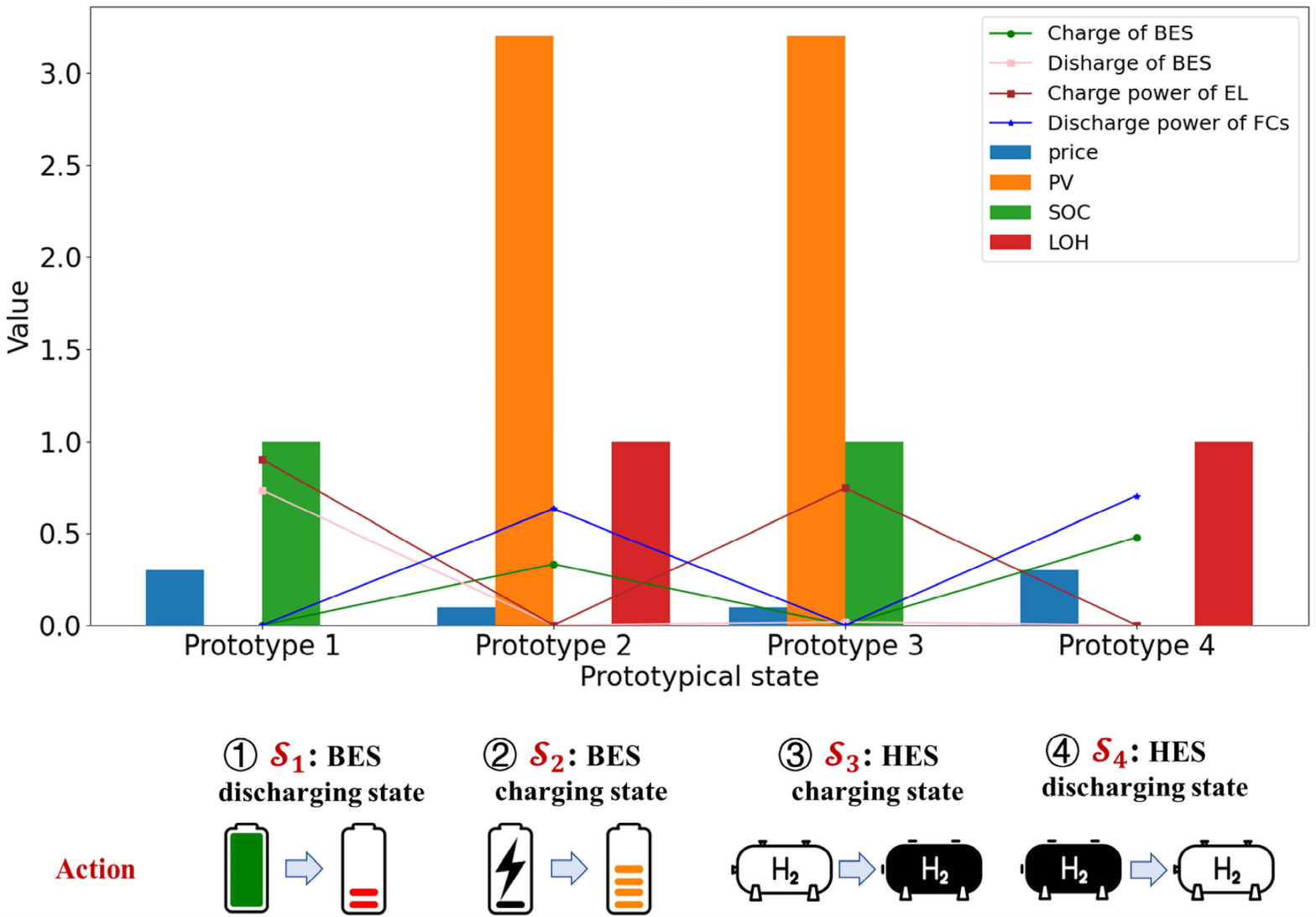}	
	\caption{The output of the prototype-based policy network for four prototypical states.}	
	\label{proto}
\end{figure}

To enhance the pre-hoc interpretability of the method proposed in this paper, we illustrate the four prototypical states and their corresponding intuitive actions in Fig. \ref{proto}. These prototypical states are designed based on common human intuition, considering the charge and discharge actions of BES and HES, which aid in elucidating the agent's policy. For instance, in Prototype 1, under conditions where the energy market price is high and PV power generation is minimal, the SoC reaches its maximum level, whereas the hydrogen storage reservoir remains empty. Drawing from common human experience, the optimal action for the BES in this scenario is to discharge and sell previously stored electricity at the elevated market price to maximize revenue. Likewise, in a scenario where the market price is lower and PV power generation is sufficient, the most advantageous action for the BES with $E_t^{\rm SoC}=0$ is to charge and store energy. This readies the system to sell electricity when prices increase and PV power generation becomes inadequate. Analogous situations also apply to the EL and FCs within the PV-ESS.

Leveraging our human-friendly prototypes, we conduct a performance evaluation comparing the proposed prototype-based policy network against the aforementioned baseline methods in the operation optimization of the heterogeneous PV-ESS. We evaluate performance using two key metrics: average reward and mean-squared error (MSE). 
The reward metric is based on the average of five trials, with each trial comprising 30 simulations. We calculate a cumulative average reward across these trials and subsequently determine the average reward and standard error over the five trials. The second metric, MSE, quantifies the dissimilarity between actions generated by the black-box model and the interpretable model during each iteration. It provides insights into how closely these methods approximate the oracle. The prototype-based policy network serves a dual purpose. First, it aims to align with the pre-trained black-box model to achieve comparable performance, and MSE serves as a tool to assess this alignment. The second purpose is to integrate human-defined prototypes, thus incorporating pre-hoc human experience. It's essential to note that the output of our prototype-based policy network is not expected to precisely replicate that of the pre-trained agent. In fact, we intentionally seek some divergence, with the goal of the network learning a new policy grounded in interpretability, informed by reasoning with prototypes and the manually specified weight matrix $W$.
\begin{figure}[htb]	
	\centering	
	\includegraphics[scale=1, width=3.5in]{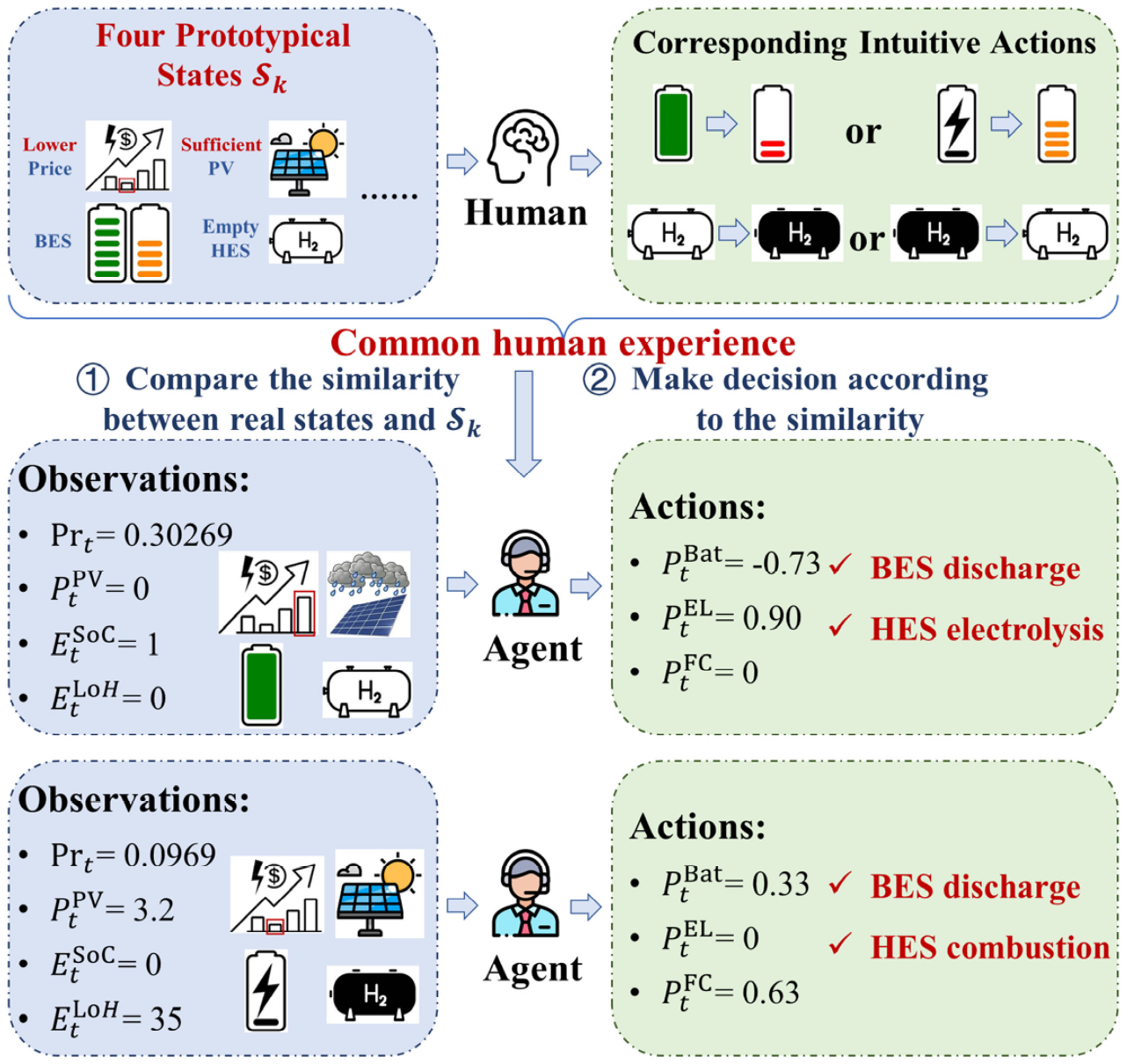}	
	\caption{The examples of interpretable decision-making by the agent.}	
	\label{example}
\end{figure}

The results are presented in Table \ref{table3} with the best-performing results highlighted in bold, and the second-best results shaded in gray. Analyzing these results, it's evident that the prototype-based policy network we introduced excels in achieving optimal performance in cases 1, 2, and 3. However, in case 4, it attains sub-optimal results in terms of reward. A comparison between the prototype-based policy network and prototype-based policy network* reveals that prototypes designed with the integration of human experience outperform learned prototypes. Having multiple prototypes proves advantageous in guiding agent selection strategies, as each can extract key information relevant to their respective actions. 

Furthermore, when compared to the K-Means method with the same number of clusters, human-designed prototypes offer more valuable insights than prototypes generated through self-classification of samples. This enriched knowledge aids in the development of superior strategies. In case 4, which exclusively involves HES, both K-Means and the prototype-based policy network achieve similar average rewards to the black-box model. Notably, the prototype-based policy network yields predictions that closely align with the black-box model, evident in the smallest MSE observed in case 4. In terms of interpretability, we also present examples of decision-making by the agent, as illustrated in Fig. \ref{example}, which unveils the correlation between the decisions made by the agent and the comprehensible decisions made by human.

\subsection{Performance comparison between different cases with different learning rate}
\begin{figure}[htb]	
	\centering	
	\includegraphics[scale=1, width=3.5in]{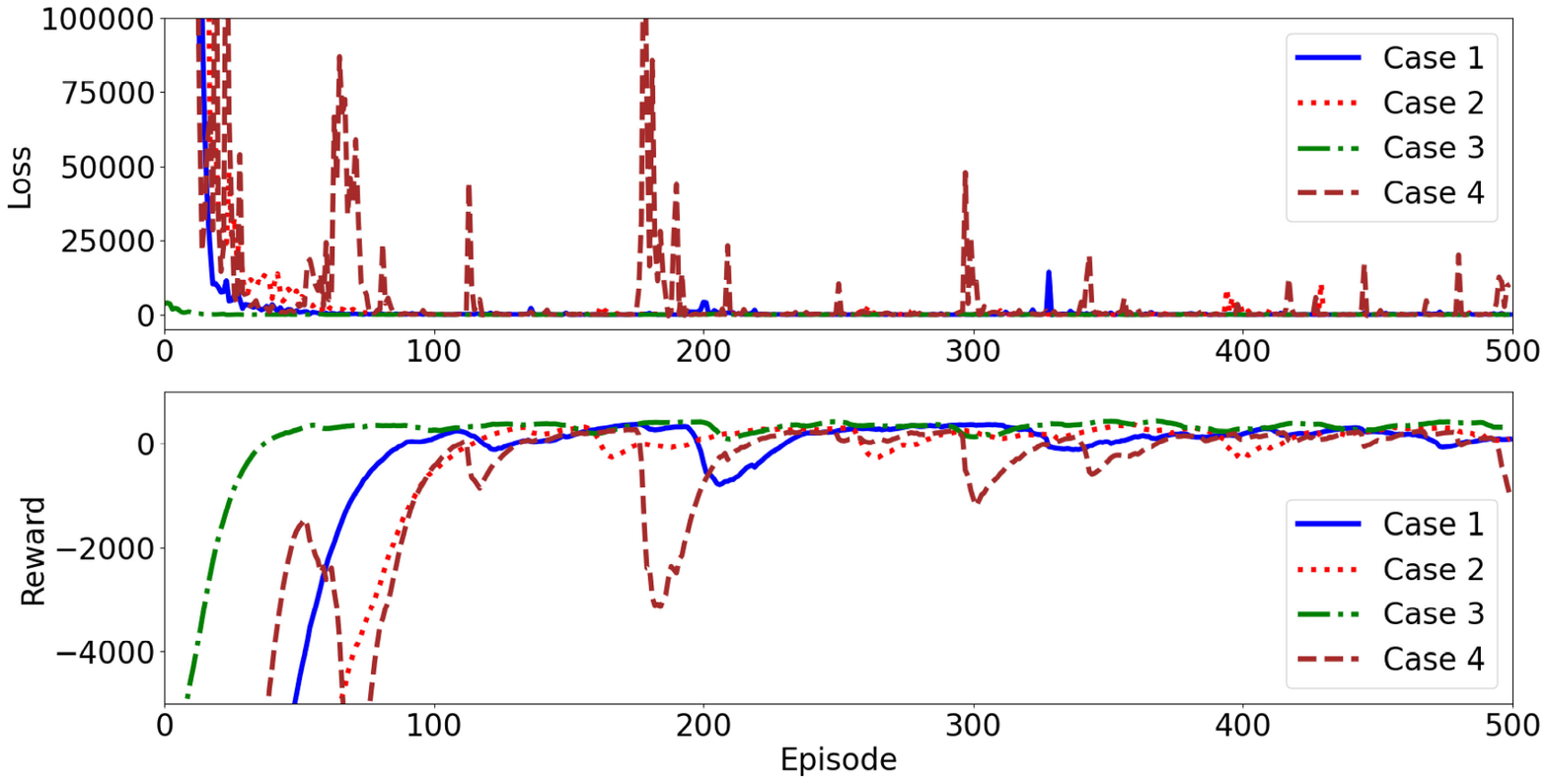}	
	\caption{Results of four different cases.}	
	\label{results}
\end{figure}

To further elucidate the influence of heterogeneity in the PV-ESS on the operator's revenue, we extend our analysis beyond the baseline comparisons and consider the performance of the pre-trained black-box model across different cases, as depicted in Fig. \ref{results}. Notably, in case 4, where only the HES is involved, the operator's profit is significantly reduced and can even result in a loss. This is primarily attributed to the fact that HES incorporates three types of equipment: the EL, the FCs, and the hydrogen storage reservoir, leading to considerably higher capital cost than other ESS. Additionally, it entails increased degradation and operation/maintenance expenses.

The noticeable increase in reward observed in case 2 compared to case 1 can be explained by the absence of any ESS-related costs in case 2. When comparing case 3 with case 1, which only accounts for the BES cost, it becomes evident that the presence of the HES significantly reduces operator's profitability.

The occurrence of negative rewards in case 4, while seldom encountered in real-world scenarios, reflects situations where users may be required to pay rental or participation fees to access the energy market with the PV-ESS. In such market, users benefit from lower-priced electricity. Additionally, the PV-ESS operator might be eligible for government incentives aimed at encouraging the use of renewable energy.

\begin{table}[thp]\footnotesize
	\centering
	\caption{Performance comparison of four cases with heterogeneous energy storage devices.} \label{table4}
	\addtolength{\tabcolsep}{4.8pt}
	\begin{tabular}{m{0.77cm} | m{0.75cm} m{0.7cm}  m{3.8cm}}
		\toprule
	Cases   & Reward & Loss & Description \\
	\midrule
	Case 1	&   151.84 & 194.28 & \scriptsize\begin{tabular}[c]{@{}l@{}}Both BES and HES are considered \\  and their costs are included.\end{tabular}     \\
		Case 2	&  182.88 & 300.13  & \scriptsize\begin{tabular}[c]{@{}l@{}}Both BES and HES are considered, \\  but their costs are not included.\end{tabular}   \\
		Case 3		&   \textbf{336.60}  & \textbf{23.01}  & \scriptsize Only BES and its cost are considered.\\
		Case 4			&   -32.17 & 1186.37 & \scriptsize Only HES and its cost are considered. \\
		\bottomrule
	\end{tabular}
\end{table}

\begin{figure}[htb]	
	\centering	
	\includegraphics[scale=1, width=3.5in]{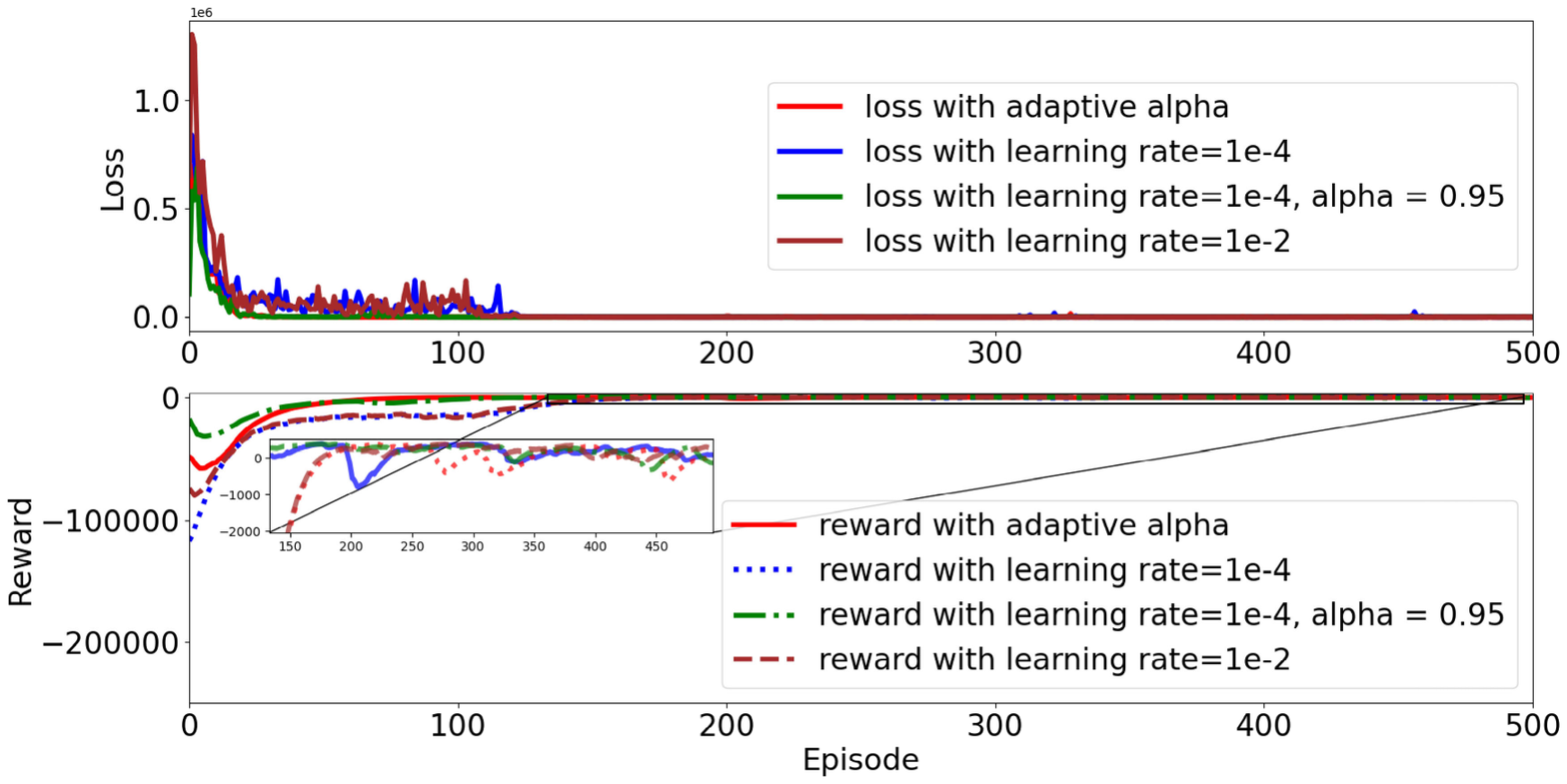}	
	\caption{Results considering different learning rates.}	
	\label{lr}
\end{figure}

The learning rate is another critical factor influencing performance. In the above experiments, we employ an adaptive learning rate, initially set at $1e^{-4}$ with an initial attenuation coefficient of $\alpha=1$. The attenuation coefficient gradually decreases with the number of simulations, following the formula $\alpha=1-\frac{{\rm step}}{{\rm total}_{\rm step}}$. To evaluate the impact of different learning rates on convergence and optimality, we design three alternative learning rate schemes: constant $1e^{-2}$, constant $1e^{-4}$, and a gradually declining learning rate with a constant attenuation coefficient of 0.95 and carry out experiments on case 1. The results are presented in Fig. \ref{lr}. As illustrated in the figure, we can observe that agents with a constant learning rate tend to exhibit slower convergence rates and are more susceptible to getting stuck in local optimization. In contrast, agents with a gradually declining learning rate demonstrate improved convergence performance. Furthermore, the adaptive attenuation coefficient proves more effective in ensuring both convergence and optimal performance compared to a constant attenuation coefficient. These findings underscore the importance of choosing an appropriate learning rate for reinforcement learning tasks.

\section{Conclusion}
In this paper, a heterogeneous PV-ESS is proposed to leverage the unique characteristics of BES and HES for scheduling tasks, with the primary objective of maximizing benefits of the PV-ESS operator through energy arbitrage. To provide more precise guidance for the operator in real-world scenarios, we present a comprehensive cost function that accounts for degradation, capital, as well as operation/maintenance costs. Additionally, in an effort to enhance the interpretability of strategies based on black-box models, we introduce a prototype-based policy network. This network utilizes human-designed prototypes to guide decision-making by comparing similarities between prototypical situations and encountered situations, leading to natural explanations of scheduling strategies. Comparative results across four distinct cases underscore the effectiveness and practicality of our proposed pre-hoc interpretable optimization method when contrasted with black-box models. Looking ahead to our future work, we plan to extend scheduling tasks to more intricate large-scale ESS featuring multiple uncertainties and heterogeneity. Furthermore, we aim to combine pre-hoc interpretable DRL with these post-hoc interpretable methods to further promote the interpretability of scheduling strategies within energy systems.

{
	\small
	\bibliographystyle{IEEEtran}
	\bibliography{IEEEabrv,mybibfile}
}





\end{document}